# Grasp-HGN: Grasping the Unexpected

MEHRSHAD ZANDIGOHAR, MALLESHAM DASARI, and GUNAR SCHIRNER, Department of Electrical and Computer Engineering, Northeastern University, USA

For transradial amputees, robotic prosthetic hands promise to regain the capability to perform daily living activities. To advance next-generation prosthetic hand control design, it is crucial to address current shortcomings in robustness to out of lab artifacts, and generalizability to new environments. Due to the fixed number of object to interact with in existing datasets, contrasted with the virtually infinite variety of objects encountered in the real world, current grasp models perform poorly on unseen objects, negatively affecting users' independence and quality of life.

To address this: (i) we define semantic projection, the ability of a model to generalize to unseen object types and show that conventional models like YOLO, despite 80% training accuracy, drop to 15% on unseen objects. (ii) we propose Grasp-LLaVA, a Grasp Vision Language Model enabling human-like reasoning to infer the suitable grasp type estimate based on the object's physical characteristics resulting in a significant 50.2% accuracy over unseen object types compared to 36.7% accuracy of an SOTA grasp estimation model.

Lastly, to bridge the performance-latency gap, we propose Hybrid Grasp Network (HGN), an edge-cloud deployment infrastructure enabling fast grasp estimation on edge and accurate cloud inference as a fail-safe, effectively expanding the latency vs. accuracy Pareto. HGN with confidence calibration (DC) enables dynamic switching between edge and cloud models, improving semantic projection accuracy by 5.6% (to 42.3%) with 3.5× speedup over the unseen object types. Over a real-world sample mix, it reaches 86% average accuracy (12.2% gain over edge-only), and 2.2× faster inference than Grasp-LLaVA alone.

CCS Concepts: • **Computing methodologies** → **Machine learning**; **Control methods**; **Computer vision**;
• **Computer systems organization** → **Robotics**.

Additional Key Words and Phrases: AI and Machine Learning, Cyber-Physical Systems

## 1 Introduction

In 2024, an estimated 2.3 million people suffered from a limbic loss and this number is projected to double by 2050 [21]. Robotic prosthetic hands provide a functional prosthetic solution for amputees. Among control methods for robotic hand actuation, fine-grained joint-by-joint control of the servos provides the most control. Yet, it requires complex control algorithms to handle multiple joints simultaneously. Moreover, acquiring user input with sufficient bandwidth and channel separation, as well as ensuring reliable real-time control is challenging. Conversely, pre-shaping to coarse-grained grasp types provides a simpler, sufficiently performant control, reducing complexity significantly.

There has been extensive work on grasp estimation to drive forward robotic prosthetic control. Approaches vary widely based on the modality of data processed. This includes visual, physiological, odometry, depth, and many more [4]. These modalities are orthogonal to each other and can be utilized in combination of each other to form a potentially better multimodal grasp estimation model. In this work, we aim to advance next generation visual grasp detection as one of the prominent sources of information.

In order for visual grasp detection approaches to perform as intended in real life, they need to be generalizable enough to new environments and object types. Every individual is unique, surrounded by a variety of objects changing over time. As shown in Figure 1, datasets can only contain a limited number of objects and it is impractical to collect a dataset that contains all world objects. Lack of generalization and robustness renders prosthetic hand impractical in real-world compared to

Authors' Contact Information: Mehrshad Zandigohar, zandi@ece.neu.edu; Mallesham Dasari, m.dasari@northeastern.edu; Gunar Schirner, schirner@ece.neu.edu, Department of Electrical and Computer Engineering, Northeastern University, Boston, MA, USA.

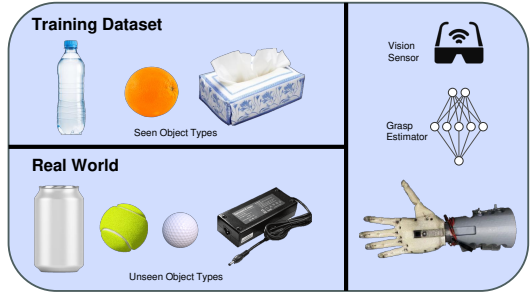

Fig. 1. Grasp Estimation for Real-World Performance.

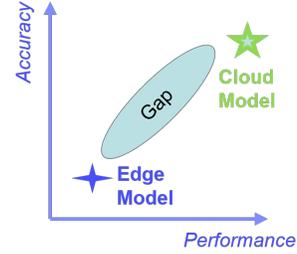

Fig. 2. The Pareto Gap.

in-lab research. Therefore, an important research question is how to construct a control method generalize to unseen object in grasp type classification.

Currently, AI methods for Visual grasp estimation [17, 26, 31, 33] have not been evaluated for real-world performance in dealing with unseen object in grasp type classification. Therefore, there is a gap in how well these models can perform on objects outside of current datasets. To truly *grasp the unexpected*, it is essential to explore the overlooked aspects of current research methodologies that are often missed in lab settings.

We define semantic projection, the ability of a model to identify the grasp type of an unseen object. As the model did not receive any guidance during training for this object type, it needs to estimate grasp type based on shared visual or semantic characteristics. We define how to quantify semantic projection and how current methods overlook this aspect. Semantically similar object types with the same grasp facilitate AI for unseen object types not present in the dataset, one step closer to real-world performance. In our experiments, we observe that conventional generalization methods e.g., image augmentation, loss regularization, and dropout do not support semantic projection. Our results indicate that a YOLO baseline as an example [17, 31] has only 15% grasp accuracy for unseen object types, close to a random classifier (7%).

Moreover, grasp estimation models are deployed on edge devices with limited compute and power budgets. In order for a robotic prosthetic hand to form the desired gesture and grasp the target object in time, it is crucial for the model to provide the correct grasp time well in advance (e.g., $> 150ms$ before contact) so there is enough time to actuate the robot. Therefore, in analyzing grasp estimation models, inference latency and the compute power needed is another deciding factor in providing a pragmatic solution. Although larger models such as VLMs can potentially increase semantic projection accuracy, due to their huge parameter space, they need to be deployed to cloud. Even when deployed to cloud, the round-trip inference latency incurred is well above the pre-shaping deadline. This creates a gap in the Pareto frontier for performance vs. accuracy trade-off as shown in Figure 2.

In this research we aim to take a step further for grasp estimation into real-world performance class by utilizing new advancements in AI. With Vision Language Models (VLMs) exhibiting superior zero-shotness capabilities due to their extensive pre-training and contextual understanding, new opportunities are possible including semantic projection. To provide a grasp detection method supporting semantic projection, we employ LLaVA [13], a VLM consisting of vision encoder and LLM. We propose Grasp-LLaVA by rethinking LLaVA to utilize the text modality for reasoning on why a grasp type is suitable for a given object. This way, we can generalize to objects unseen in the training dataset using semantically similar object types seen during training.

In deploying Grasp-LLaVA, the parameter size is humongous and computational demand of such LLM-based models are high, creating a Pareto gap in performance and semantic projection accuracy (Figure 2). To this end, we propose hybrid grasp network (HGN), an edge-cloud infrastructure by deploying Grasp-LLaVA as a universal model to the cloud while deploying a well-calibrated specialized model on edge, effectively balancing the accuracy and performance and expanding the Pareto front.

To summarize, the contributions of this papers are:

- Definition of Semantic Projection, how to quantify it and how well current methods exhibit semantic projection.
- Grasp-LLaVA, a grasp detection method supporting semantic projection by incorporating grasp reasoning.
- HGN, a deployment infrastructure with a fast edge specialized model and a cloud universal model having high semantic projection generalization; balancing accuracy and performance trade-off.

Our results demonstrate that Grasp-LLaVA achieves 50.2% accuracy in semantic projection, outperforming other methods by 13.5%-34.9%, one step closer to real-world performance. Moreover, we demonstrate that HGN contributes to the latency vs. accuracy trade-off by expanding the Pareto front and fill the Pareto Gap, with calibrated models providing better fronts. As an example, on an NVIDIA Jetson Orin NX 16GB, HGN can increase the semantic projection accuracy to 45% (+8.2% over the baseline edge model) while maintaining the average latency below a 150*ms* pre-shaping deadline. Lastly, we evaluate HGN from the user perspective and provide User Upsetness Index (UUI), effectively measuring the correctness and lateness of each method and their impact on the user experience. HGN with confidence calibration (DC) enables dynamic switching between edge and cloud models, improving semantic projection accuracy by 5.6% (to 42.3%) with 3.5× speedup over the unseen object types. Over a real-world sample mix, it reaches 86% average accuracy (12.2% gain over edge-only), and 2.2× faster inference than Grasp-LLaVA alone.

## 2 Related Work

Current visual grasp estimation approaches can broadly be categorized into classification and detection based approaches as follows:

*Classification-Based Approaches.* Most works in visual grasp estimation are classification-based approaches. In these works, often a pretrained popular model from the ImageNet family is fine-tuned over a custom dataset used for grasp classification. [22] utilizes AlexNet to classify 4 grasp types. In [18], GoogleNet and AlexNet are trained to classify 5 grasp classes. [30] uses InceptionV3 architecture to provide a probabilistic estimation of 5 grasp types. [5] utilizes VGG-16 and a Fully-Connected (FC) layer for the same purpose. In [33], the authors utilize a pretrained MobileNetV2 to extract visual features, and use a Bayesian Multi-Layer Perceptron (MLP) to form a grasp probability. Similarly, [26] exploits MobileNetV2 for feature extraction accompanied by a FC layer to pre-shape grasp for 15 objects.

Although the aforementioned approaches can achieve high accuracies in grasp classification, they are trained and evaluated on cropped images that only contain the desired target object. This is far from ideal as in most cases with a prosthetic hand, not only the field of view contains multiple objects but also the target object occupies only a small portion of the view. This makes detection models more favorable as they detect the bounding box with the object and provide the grasp type for that target object. It is imperative to note that none of the aforementioned studies analyze model performance on unseen object types.

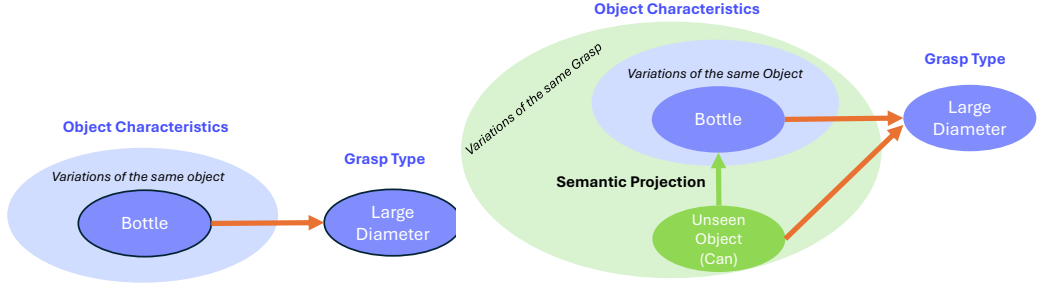

Fig. 3. Venn diagram showing conventional generalization.

Fig. 4. Venn diagram showing semantic projection.

*Detection-Based Approaches.* Taking one step closer to real-world performance, detection-based models process images with a natural field of view containing multiple objects within their natural scale. In doing so, YOLO has been the most widely used detection-based approach for many applications including grasp detection due to its high accuracy and lower latency. [17, 31] utilize YOLOv4 and YOLOv5 to detect the desired target object with the help of user gaze. Although the aforementioned approaches are more effective and realistic compared to classification-based approaches, they still lack performance analysis on unseen object types.

Since the related work for grasp estimation has not caught up with the recent advancements in AI yet, there is an opportunity to exploit them. Therefore, to evaluate detection-based models in our experiments, we utilize a more recent YOLO version [27]. Moreover, to enhance our classification-based baseline, we employ ViT [2] and CLIP [20] as the most advanced classification models available, similar to [32].

## 3 Semantic Projection

Conventional generalization methods such as image augmentation [23], loss function regularization, weight decay [6], and architectural modification (e.g., dropout [7]) have been used for many years to improve generalization of models [23]. These methods have shown to help reliably increase model accuracy in detecting variations in the same object class. This can be showed using a Venn diagram (Figure 3) for grasp estimation. As an example, the generalization of a model trained on images of a bottle can be further increased by utilizing conventional generalization techniques to help reliably detect the same object in different settings.

However, the problem of grasp detection (Figure 1) is more complex. Current grasp estimators are trained on a dataset with limited number of objects. Yet, there are countless object types in the real world that will not be present in the training dataset. Among this virtually infinite number of object types, many of them can be grasped with the same grasp type to those seen in the training dataset. However, this doesn't mean that training a model on a limited set of object types with those grasp types will provide a sufficiently generalized model to detect any of the objects with the same grasp type.

In fact, our results on conventional models reveals that such phenomenon does not happen. As an example, a fully trained YOLO model capable of detecting grasp types for seen object types with high accuracy of 80.3% fails miserably with 15.3% semantic projection accuracy (Figure 13) when given objects not seen by the model during training - only 7% higher than a random classifier. This shortcoming is especially more unexpected given that the grasp type used for the aforementioned unseen object types have been seen during training through other object types.

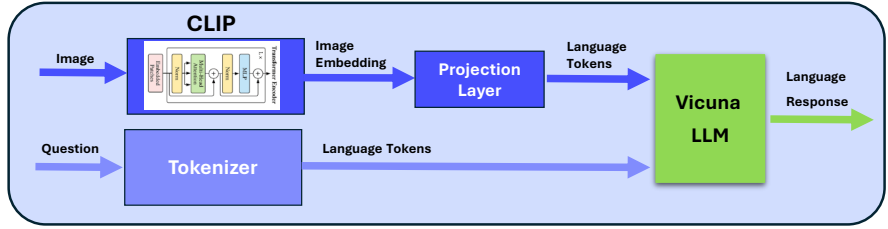

Fig. 5. LLaVA design overview and components.

To define such relationship between a given grasp type and object types belonging to it, we introduce semantic projection (Figure 4). Unlike traditional generalization, which typically relies on explicit training data for new object types, semantic projection leverages shared visual and semantic characteristics to infer grasp types for unseen objects. Object types with the same visual characteristics can most likely be projected to the same grasp type. Consequently, a model capable of semantic projection will be able to correctly predict the grasp type for unseen object types. As an example, a soda can will likely be semantically projected to Large diameter grasp type given that it is semantically similar to a water bottle, with both having the same shape, size and diameter.

To evaluate the real world performance of the model, we assess its semantic projection. To this end, a given dataset is split into seen and unseen sets by their object types. However, unlike conventional data splitting, it is not enough to simply divide the samples into seen and unseen images. For assessing the semantic projection, it is also essential to identify the objects within the images and distinguish between seen and unseen object types. The training and validation sets therefore only include samples from the seen set, while unseen object type samples are reserved solely for the test set.

In our experiments, we will evaluate each method against their conventional accuracy, reporting the grasp detection accuracy on seen object classes, along with semantic projection accuracy.

## 4 Grasp-LLaVA

The rise of generative AI has provided the research community with a vast opportunities not seen before. Models such as ChatGPT, Stable Diffusion, DALL-E have shown capabilities never imagined before and they have been extensively used in many areas [9, 12, 16, 24]. Since such models have been pretrained on internet-scale data, and their ability to reason among other, they have been shown to be good zero-shot learners. Given this, We take advantage of Vision Language Models (VLM) to enable grasp estimation in cases with unseen objects resulting in better generalizability.

To implement Grasp-LLaVA we use a general-purpose VLM called LLaVA (Large Language-and-Vision Assistant) [13]. LLaVA is a well-known general-purpose VLM with remarkable accuracy. The model's architecture (as shown in Figure 5) enables it to process and reason about multimodal inputs, providing a more holistic understanding and interaction with complex data.

To train Grasp-LLaVA, Figure 6 outlines the steps required. We first begin by describing the dataset Mix consisting of MDS-1 and HANDSv2. Next, the Object-Aware Data Splitter aims at grouping the object classes into seen and unseen categories based on their grasp type. Afterwards, the conversation Generator generates the ground-truth textual labels used during Grasp-LLaVA's training. Finally, with the data ready, LLaVA Trainer the pretrained LLaVA model over our processed training set. To compare Grasp-LLaVA's accuracy against Grasp-ViT and Grasp-CLIP, all methods are evaluate in detail in subsection 6.3.

### 4.1 Datasets

Among the visual datasets publicly available online, two datasets stand out in the area of grasp detection for prosthetic hand control: HANDSv2 [31] and MDS-1 from MeganePro [1]. HANDSv2 consists of 530 videos with each one averaging about two minutes which consist of 53 object classes. We sampled 10000 images using uniform distribution from the videos to have a dataset of images.

MDS-1 [1] consists of 45 videos with each averaging around one hour constituting 18 object classes. However, its ground truth data of bounding boxes is corrupted. Manually labeling all 10000 frames to create the ground truth is too labor intense. Therefore, we regenerated object and gaze labels using an overfitted YOLO model (only to generate ground truth). To this end, we trained a YOLO model on a small labeled subset and deliberately overfit it. Overfitting is desired as the model annotates that same subset on visually similar frames. This follows standard pseudo-labeling practice, where confident predictions from a trained model are used to create labels on unlabeled data [11].

Similar to HANDSv2, 10000 images were sampled with uniform distribution over the labeled data to constitute our imagery dataset. Moreover, among the common object classes from the two datasets, the grasp type for the ones from MDS-1 that had a different grasp type for the same object were replaced by their corresponding grasp type from HANDSv2 to stay consistent.

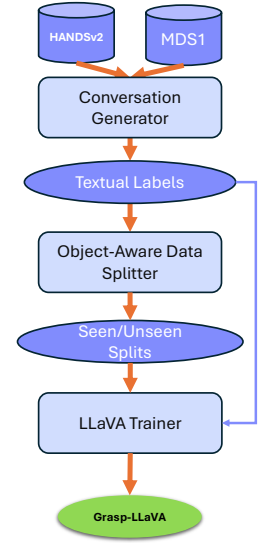

Fig. 6. Overview of Grasp-LLaVA Training Pipeline.

To eliminate the effects of domain-shift and particularly focus our study on the generalizability capabilities of each model, instead of picking only one of the aforementioned datasets, we approached our data preparation by mixing both datasets. This dataset mix will eliminate other factors impacting the accuracy evaluations and only focuses on the generalizability to unseen objects.

*Enhancing Accuracy Through Focus Points.* Given many of the grasp estimation models are purely classification model, and that the field of view can contain several objects, it is imperative to provide a means for the model to focus on the target object. To this end, we utilize the gaze point to provide the object of interest, providing the coordinates of the user's gaze. Contrary to detection models such as YOLO where the focus point is directly utilized to select the correct bounding box, for classification based methods we indicate the focus point by superimposing each image with a distinctive green star.

### 4.2 Object-Aware Data Splitter

The purpose of the Object-Aware Data Splitter is to evaluate how models generalize. The Object-Aware Data Splitter Deliberately separates object classes and categorizes them into Seen or Unseen categories. Seen object classes will be used during training (34 object classes) classes and Unseen object classes will be used in evaluation only (28 object classes) to emulate objects that are new in the real world but not in the dataset. This provides a suitable benchmark for evaluating semantic projection.

Furthermore, we split the seen set into two sets, train set (90%) and validation set (10%). This way, we monitor how well the training goes without involving any of the data from the unseen category.

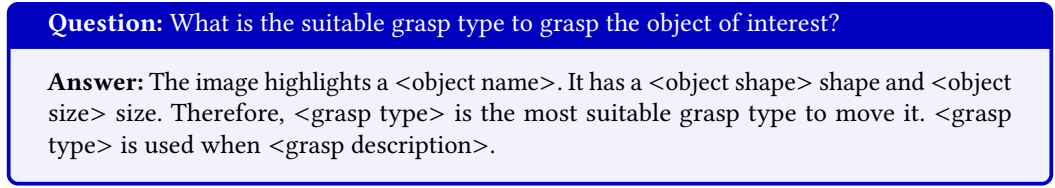

Fig. 7. Template used for Grasp Reasoning During Training.

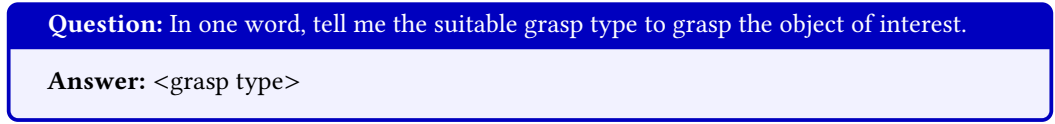

Fig. 8. Template used During Inference.

### 4.3 Conversation Generator

Oftentimes, deep learning models in vision research are trained over a set of images and their corresponding ground truth. This ground truth is usually the class id in case of classification tasks and the coordinates for the bounding box for object detection on top of the class id. However, for VLMs the ground truth is pairs of images and their corresponding textual conversations.

In our Grasp-LLaVA case, we utilize these conversations to generates the textual labels that help inject reasoning into why a grasp type is suitable. During the training phase of Grasp-LLaVA, each iteration involves feeding the model an image and its corresponding conversation (Figure 7). The conversation consists of only one question and its answer (hence not back and forth). The question is: "What is the suitable grasp type to grasp the object of interest?". The answer is a detailed sentence combining the object's name, size, shape, target grasp type, and the reasoning behind why that grasp type is suitable for the object. This approach teaches the model to associate object characteristics (e.g., size and shape) with the selection of the correct grasp type, emphasizing the reasoning process. As a result, Grasp-LLaVA is fine-tuned to function as a question/answer assistant specialized in identifying suitable grasp types by considering object-specific attributes.

During inference (Figure 8), the trained model is provided with an input image and asked one simple question: "In one word, tell me the suitable grasp type to grasp the object of interest". At this stage, no additional hints, such as object size or shape, are provided. Since the model has been trained to reason about why a particular grasp type is appropriate, it is capable of identifying a grasp type even for objects that were not present in the training or validation sets.

This reasoning aligns closely with the concept of semantic projection, as it connects object-specific features such as object name, shape, and the correct grasp type to the rationale behind the suitability of that grasp type. By establishing these semantic connections, the model leverages shared features to generalize grasp selection across object types. The object reasoning for each grasp type and their corresponding shape and size have been utilized from [14], which proposes a taxonomy of grasps based on how objects are used in everyday activities.

| Parameter | Value |
|---|---|
| Lora R | 128 |
| Lora Alpha | 256 |
| MM Projector LR | 2.00E-05 |
| Num Train Epochs | 1 |
| Train Batch Size | 16 |
| Learning Rate | 2E-04 |
| Warmup Ratio | 0.03 |
| Lr Scheduler Type | cosine |
| Model Max Length | 1024 |

Table 1. Hyper-parameters used during training Grasp-LLaVA.

### 4.4 LLaVA Trainer

The purpose of LLaVA trainer is to fine-tune LLaVA on the training set. Since training LLaVA from scratch requires tremendous resources and time, and that our dataset is not as huge as the data used for training universal VLMs, we will take advantage of LoRA [8] for fine-tuning. LoRA (Low-Rank Adaptation) reduces the number of trainable parameters in large language models by freezing pre-trained model weights and injecting trainable low-rank matrices into each layer of the Transformer architecture. This method significantly lowers the computational and memory requirements, making training more efficient. By only training these smaller matrices, LoRA achieves high training throughput and model quality comparable to full fine-tuning, while also reducing GPU memory usage and storage requirements, thereby accelerating the training process. To fine-tune LLaVA over our dataset, we use the hyper-parameters outlined in Table 1.

## 5 Hybrid Grasp Network (HGN)

Although our vision language model (Grasp-LLaVA) show generalizability to unseen object classes, deploying such models to embedded devices is challenging. Most often, such models are incredibly compute-intensive and possess billions of parameters with Grasp-LLaVA consisting of over 7B parameters. Given the limited resources on edge devices, deploying the aforementioned models on edge is virtually impractical. An example of this is deploying LLaVA on NVIDIA's Jetson Orin NX 16GB. This edge device is among the most advanced edge devices with 275 trillion operations per second (TOPS). However, even in super mode, Orin can only produce 0.6 tokens/sec [15].

With the current limitations of edge devices in executing advanced generative models, and the exponential growth of such models in size, a paradigm shift in system infrastructure design is necessary. As such, this infrastructure needs to employ a specialized model on edge for faster inference while a universal model is deployed on cloud as a safeguard.

To this end, we propose Hybrid Grasp Network (HGN). In the next subsections, we first outline the components that constitute HGN. Then, we discuss how to adapt specialized models for integration into HGN. Finally, we provide insights into how to tune HGN based on performance and accuracy requirements.

### 5.1 HGN Infrastructure

To address the issue of inference latency for VLMs we propose HGN, a deployment infrastructure consisting of an edge specialized model with desirable latency but low generalization; and a cloud-based universal model with desirable generalization being offloaded to the cloud. This is depicted in Figure 9. Each component is utilized as follows:

*5.1.1 Cloud Universal Model.* Given the poor semantic projection of specialized models, a universal model utilizing VLMs such as Grasp-LLaVA is necessary to enable accurate predictions for unseen object types. However, the impressive accuracy of such universal models comes at a cost. A universal model is orders of magnitude larger than specialized models. With the current trends in advancements in VLMs, they are expected to get even larger with models over 100s of billions of parameters. Therefore, it is imperative to deploy them on a cloud server machine with highly performant GPUs. This translates into longer inference time and added network round-trip delays.

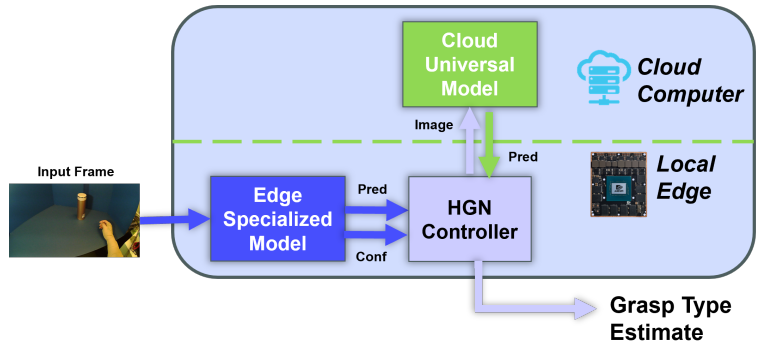

Fig. 9. HGN design overview and components.

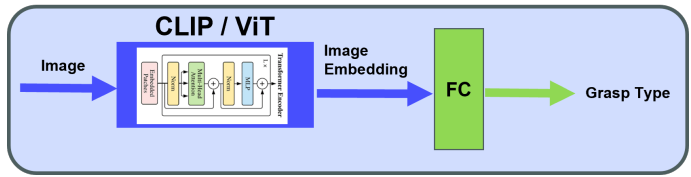

Fig. 10. Grasp-ViT and Grasp-CLIP Architecture.

*5.1.2 Edge Specialized Model.* Although we observe that conventional classification- and detection-based models do not exhibit semantic projection compared to their VLM counterparts, they are still sufficiently accurate for seen object types and can occasionally predict unseen object types accurately, which makes them a suitable specialized model.

To this end, we have constructed and trained a suite of specialized models tailored for the edge, namely Grasp-YOLO, Grasp-ViT, and Grasp-CLIP. With Grasp-YOLO, we fine-tuned YOLOv8 for grasp detection with minimal architectural changes. Our Grasp-YOLO utilizes a more recent version compared to SOTA in detection-based approaches. Moreover, we significantly enhance the classification-based approach by designing Grasp-ViT and Grasp-CLIP (Figure 10). Grasp-ViT employs a ViT backbone, followed by a fully connected classifier head composed of three layers with 768, 256, and 13 neurons, each with ReLU activations. Similarly, Grasp-CLIP builds on the pretrained CLIP vision backbone - the same backbone used in Grasp-LLaVA - followed by a similar custom head for grasp classification.

Due to their parameter size, our edge model suite are almost two orders of magnitude faster than VLMs. Given their relatively small size, the specialized model can be easily deployed to edge devices. Such deployment enables fast inference locally and avoids computation on cloud.

Although we evaluate the performance of each model, we choose Grasp-CLIP as our edge specialized model as it is among the most advanced image classification models. Moreover, we strategically select Grasp-CLIP for our experiments since CLIP also acts as the vision encoder for Grasp-LLaVA. Hence, it provides more insights as to how well the addition of the grasp reasoning assist in semantic projection.

While we anticipate the edge specialized model provides a fast and somewhat satisfactory grasp prediction, it yet suffers from poor generalization, especially in real-world use-cases. The edge model uses a calibrated confidence score, obtained via temperature scaling or density-aware calibration, to decide whether to accept its prediction or offload to the cloud.

*5.1.3 HGN Controller.* HGN controller aims to balance between the fast inference time of the edge specialized model and the higher accuracy of cloud universal model. HGN controller continuously monitors the prediction confidence of the edge specialized model and makes the ultimate decision to use specialized model's prediction as the selected grasp type or to incorporate the cloud universal model as a fallback mechanism. This means that if the prediction confidence of the edge specialized mode is lower than a threshold, HGN controller assumes that as a misprediction and involves the cloud universal model to predict the grasp type. This mechanism ensures that the cloud universal model is only utilized when the edge specialized model is performing poorly, hence improving latency dynamically while mitigating accuracy drops.

Given that the decision to alternate between each model is based on the prediciton confidence of the specialized model, it is extremely important for the confidence score to be realistic. In the next section, we discuss how well the confidence score represents reality.

## 5.2 Prediction Confidence

In real-world applications, a model must not only be accurate, but also should indicate when they are likely to be incorrect. Prediction confidence therefore serves as an indication of how likely it is for a prediction class to reflect the real ground truth. This is a crucial aspect of the edge specialized model since it provides the confidence score that is used by HGN Controller to decide whether to engage the universal model during inference.

In neural networks, the output logits are converted into probabilities using the Softmax function. The confidence score ($S(x)$) of a prediction is then simply the highest probability assigned to the class with the highest probability:

$$S(x) = max_c P(y = c|x) \tag{1}$$

where $x$ is the input frame to the specialized model and $c$ is the grasp type with the highest probability. However, this doesn't necessarily mean confidence score correctly represents the ground truth likelihood of a grasp type being the correct grasp. In the remainder of this subsection, we evaluate confidence correctness to identify if this assumption holds true and discuss potential approaches to tackle this issue.

*5.2.1 Confidence Correctness.* Confidence correctness indicates how well the probability associated with the predicted class label reflects its ground truth correctness likelihood. This means that if the confidence score of a given model for 100 input samples is 0.4, and the model's confidences are all correct, the model can only classify 40 samples correctly. This is also known as calibration of a model. A well-calibrated model is neither over- or under-confident. A model is over-confident if the predicted confidence is systematically higher than the actual correctness probability. A model is under-confident if the predicted confidence is systematically lower than the actual correctness probability.

To evaluate confidence correctness, there are generally two approaches: reliability diagrams and Expected Calibration Error (ECE) [3]. Reliability diagram, a.k.a. calibration plot, is a visual approach where the x-axis is the predicted confidence, and the y-axis is the actual accuracy. Here, the dataset is divided into $K$ bins based on confidence intervals. Each bin $B_k$ contains all predictions where the confidence falls within that range. The average confidence for each bin will then be:

$$conf(B_k) = \frac{1}{|B_k|} \sum_{i \in B_k} S(x_i) \tag{2}$$

While the actual accuracy for the bin is:

$$acc(B_k) = \frac{1}{|B_k|} \sum_{i \in B_k} 1(\hat{y}_i = y_i) \tag{3}$$

where 1 refers to the indicator function.

Based on this, a perfectly calibrated model lies on the diagonal line where confidence matches accuracy. However, an over-confident model will have a curve below the diagonal, while in an under-confident model the curve is above the diagonal.

Expected Calibration Error (ECE) provides a scalar summary statistics of reliability. Similar to reliability diagrams, ECE quantifies the difference between confidence and actual accuracy over all

bins:

$$ECE = \sum_{k=1}^{K} \frac{|B_k|}{n} |acc(B_k) - conf(B_k)| \quad (4)$$

Here, A low ECE means the model is well-calibrated. An acceptable ECE is often $ECE < 0.05$. The reliability diagram in Figure 16a, gives us insights into Grasp-CLIP's confidence correctness. As evident, the model is over-confident throughout all probability intervals. Moreover, the ECE value is well beyond the desired ECE.

In order for the HGN controller to function properly, it is important that the specialized model produces well-calibrated confidence estimates. Hence, to address over-confidence of the specialized model, confidence calibration is necessary.

*5.2.2 Confidence Calibration.* There are many approaches to calibrate confidence of a model. Most common approaches include Histogram binning [28], Isotonic Regression [29], Platt Scaling [19], Dirichlet Calibration (DC) [10], Temperature Scaling (TS) [3] and Density-Aware Calibration (DAC) [25]. We focus our efforts on DC, TS and DAC since the other methods are a variations of these approaches. DC is multi-class probability calibration method that extends Platt scaling by applying a Dirichlet-based transformation for improved reliability. TS is a simple post-processing method that rescales logits using a single temperature parameter to produce better-calibrated probabilities. DCA is a probability calibration approach that adjusts predictions based on data density, improving reliability in underrepresented regions.

As part of our experimental results we will investigate the benefits of calibration, and evaluate their impact on the overall system's performance.

## 6 Experimental Setup and Results

To set up and evaluate the proposed method against semantic projection, the experimental setup and results for SOTA and Grasp-LLaVA are discussed in detail. Moreover, we explore calibration methods, evaluate the HGN controller based on the calibrated edge specialized model and finally demonstrate the effectiveness of HGN in expanding the performance vs. accuracy trade-off.

### 6.1 Experimental Setup

For conventional generalization, we use center cropping, flipping, and color hue as image augmentation [6, 7, 23, 23] during training for all models and utilize dropout and label smoothing as structural techniques for generalization.

During training, all edge specialized models were fine-tuned for 400 epochs with a batch size of 64, using the ADAM optimizer and Cosine Annealing learning rate scheduling.

For deployment on an embedded device, NVIDIA Jetson Orin NX 16GB would be a suitable option due to its high computational capabilities yet portable design. Jetson Orin NX reaches 100 TOPS and benefits from a 1024-core NVIDIA Ampere architecture GPU with 32 Tensor Cores. Each grasp estimator can be further optimized in latency with NVIDIA's hardware-specific optimizations.

### 6.2 Enhancing Accuracy Through Focus Points

Most often, there will be multiple objects present in the visual field of view, and therefore, it is important to identify the object of interest. This is achieved by using an eye-tracker's gaze data to identify the point of focus, and thereby selecting the user's object of interest to be grasped.

In detection-based approaches, the coordinates of the focus point is used directly to find the closest bounding box of an object and deduce the target object. However, this is not possible in classification-based approaches where the model only yields the class probabilities and no bounding boxes. To this end, in order to indicate the object of interest in the input image for classification-based models, we highlight the focus point in the input image with a green star. This is shown in Figure 11.

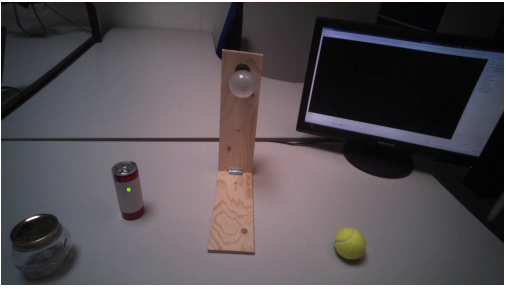

Fig. 11. Example of incorporating a focus point. Note the green star superimposed on the can.

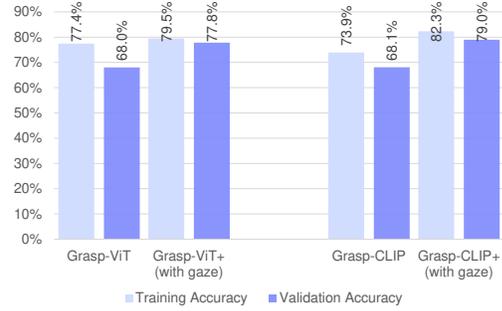

Fig. 12. Accuracy for Grasp-Vit and Grasp-CLIP (Without gaze) vs. Grasp-Vit+ and Grasp-CLIP+ (with gaze) @200 epochs.

As observed in Figure 12, without providing the focus point, although the training accuracy reaches its peak, the validation accuracy is stuck at 68%. This is due to the fact that the model is prone to overfitting when the focus point is not provided. On the other hand, when the focus point is provided, validation accuracy can reach its expected accuracy with an increased average accuracy of 10.35%. Given this crucial role of focus point in grasp detection, we have incorporated gaze points in all our analysis.

### 6.3 Evaluating Semantic Projection

With grasp estimation models fully trained on the training set, we evaluate each model's generalizability to unseen object types. In doing so, it is common to have a random chance baseline to asses how well each method performs compared to a random chance classifier (RND). Therefore, given that we are modeling 13 grasp types, accuracy results greater than $\frac{1}{13} = 7.67\%$ are better than random chance classifier.

Evaluation results for conventional generalization is demonstrated in Figure 13. We observe that all models perform well and meet expectations. Since the difference between training and validation accuracy is minimal, we can deduce that no overfitting has occurred. Given the high accuracy of each model, they are in line with the desired in-lab results as seen in grasp estimation research.

On the other hand, when evaluating results for semantic projection we observe that Grasp-YOLO performs extremely poorly when faced with unseen object types. In fact, Grasp-YOLO is performing only 7% better than a random chance classifier when faced with unseen object types. This is because YOLO is primarily designed for object detection, focusing on speed and localization accuracy rather than understanding semantic relationships or global context. Grasp-CLIP+ and Grasp-ViT+, however, perform slightly better due to the use of self-attention mechanisms, enabling it to capture long-range dependencies and relationships across the entire image. We can observe that Grasp-CLIP+ outperforms Grasp-ViT+ potentially due to its joint training with language modality. Nonetheless, this improvement is yet suboptimal. Grasp-LLaVA on the other hand, outperforms all other models with an impressive semantic projection accuracy of 50.2%, showing the superiority of such reasoning framework over current grasp estimation models.

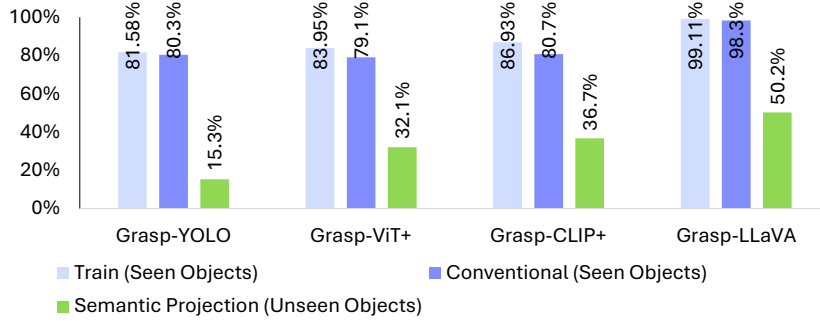

Fig. 13. Conventional and semantic projection accuracy.

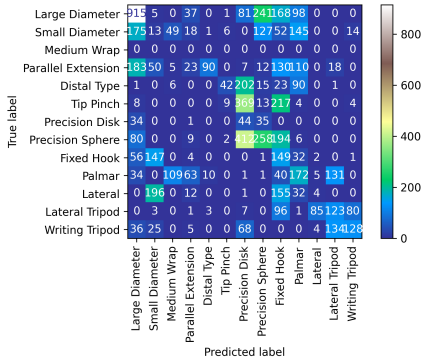

Fig. 14. Grasp-YOLO Confusion Matrix.

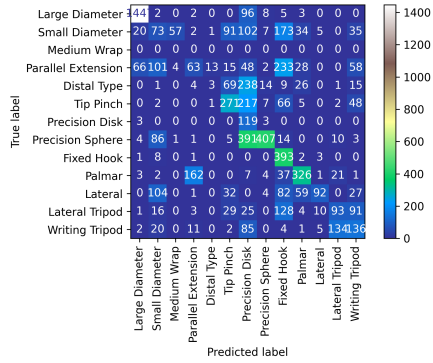

Fig. 15. Grasp-LLaVA Confusion Matrix

The confusion matrices demonstrate a clear performance distinction between Grasp-YOLO (Figure 14) and Grasp-LLaVA (Figure 15). Grasp-LLaVA significantly enhances semantic projection capabilities, achieving notably higher accuracy on unseen object types and substantially reducing misclassifications between visually similar grasp categories, particularly between Precision Sphere and Tip Pinch. This improvement is evidenced by the stronger diagonal prominence and higher correct classification counts observed in the Grasp-LLaVA confusion matrix, underscoring its superior ability to generalize beyond the training set.

However, both models continue to struggle with specific categories, notably Medium Wrap and tripod-based grasps like Lateral Tripod and Writing Tripod. These persistent difficulties likely arise from limited visual distinctiveness or insufficient representation within the training datasets. Such findings underscore the importance of dataset refinement, including increased representation and clearer definition of ambiguous grasp types.

Despite these challenges, both YOLO and Grasp-LLaVA perform well on common grasp types such as Parallel Extension and Precision Sphere. While YOLO maintains reasonable accuracy on seen object types, the improvements offered by Grasp-LLaVA—particularly in handling unseen objects—highlight its clear advantage and make it a more promising solution for real-world prosthetic grasp control.

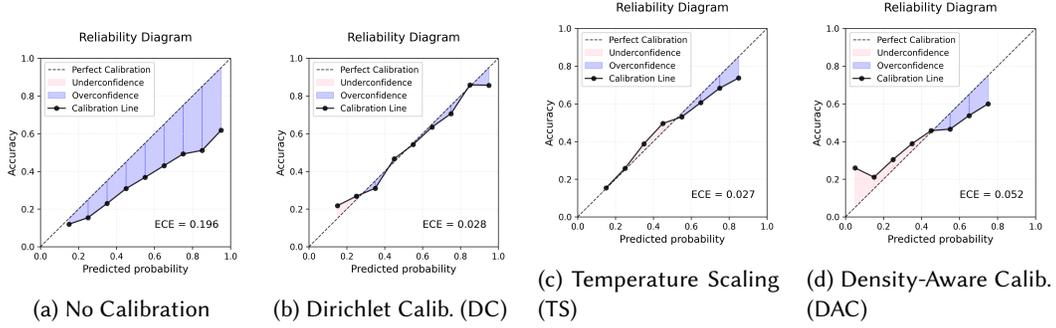

Fig. 16. Reliability diagrams before and after applying different calibration methods.

## 6.4 Evaluating HGN Controller

We calibrated Grasp-CLIP+ with DC, TS, and DAC methods over 10 bins as is the most frequently used configuration in literature and enables a more intuitive understanding. The reliability diagrams and their corresponding ECE are demonstrated in Figure 16.

We can observe that based on the reliability and ECE results, the confidence correctness has significantly improved. All ECEs are almost within < 0.05 with TS having the least ECE. Although DAC still has an acceptable ECE, we can observe that it results in a model that is under-confident for probabilities below 0.5 and over-confident for the ones above. This is more alleviated in TS and DC where the calibration line almost matches the perfect calibration line.

After calibration, the HGN controller can incorporate reliable confidence scores into its decision making. The ideal model selection used as HGN's ground truth is outlined in Table 2. Here, if the edge model ($E$) has a correct grasp estimation, edge model is selected as the ideal model. This is independent of how the cloud model ($C$) performs as the edge model is both correct and the faster choice. However, when the edge model is incorrect, when the cloud model is correct, the ideal selection is the cloud model. In case both models are incorrect, any of the models can be selected. For model selection when latency is crucial, the edge model is favored due to it's lower latency.

To evaluate the effectiveness of HGN, it is crucial to not rely on a simple accuracy metric. Since the goal of HGN controller is to involve the universal model only if the edge model's confidence is lower than a specified threshold, we define the positive class as "keep on edge" and the negative class as "move to cloud". Based on the ideal model selection outline in Table 2, a true positive then can be specified as HGN's decision to keep computations on edge while the ideal model is edge, which will result in speedup gains. On the other hand, false positive is HGN's decision to keep on edge while ideally it should have been the cloud model, resulting in accuracy degradation. A true negative is then HGN's decision to move to cloud while being the ideal model, resulting in accuracy improvements. Conversely, a false negative is HGN's decision to move computations to cloud while edge model is the ideal choice, therefore harming the latency or even accuracy.

Table 2. Ideal Model Selection for Ground Truth.

|  |  | *Cloud* | |
|---|---|---|---|
|  |  | *Inorrect* | *Correct* |
| *Edge* | *Inorrect* | E | C |
|  | *Correct* | E | E |

Table 3. Classification Outcomes.

|  |  | *GT* | |
|---|---|---|---|
|  |  | E | C |
| *Pred* | E | TP | FP |
|  | C | FN | TN |

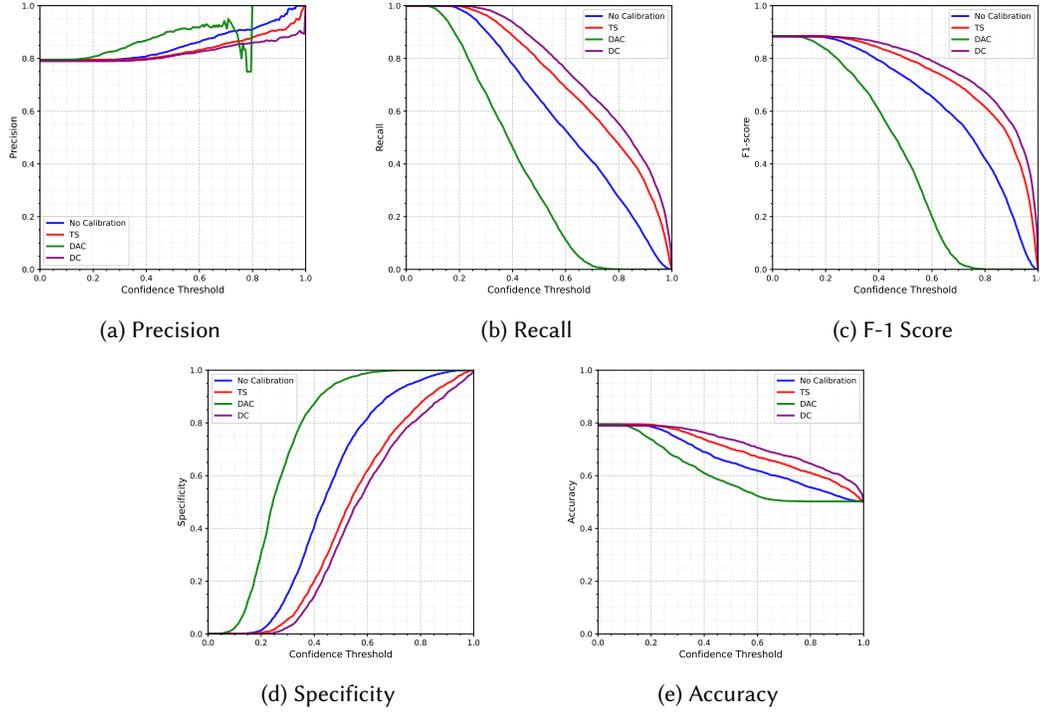

Fig. 17. Classification metrics for different calibration methods.

Based on these definitions, precision, recall, F1-score, specificity and accuracy of HGN for different thresholds can be calculated (Figure 17). Each metric therefore can be interpreted as follows:

*Precision.* depicted in Figure 17a, indicates that out of all cases where the model decided to employ the edge model, how many were the ideal choice. High precision ensures that when the system keeps a task on edge, it is actually beneficial for speedup. On the other hand, low precision means that the system wrongly employs the edge model, hence hurting accuracy. Given that higher thresholds exclude predictions with lower confidences, precision increases when HGN threshold is higher.

*Recall.* Indicates that out of all cases where choosing the edge model was the ideal choice, how many did HGN correctly decide choose. High recall ensures that when keeping on edge is ideal, the system actually makes that choice. On the other hand, low recall means that many cases that could be kept on edge for speedup are instead sent to the cloud, which adversely affects latency. Since lower thresholds favor the edge specialized model's predictions, recall is highest when threshold is at its lowest. This is also observed in Figure 17b.

*F-1 Score.* Measures a balance between precision and recall. It indicates how well the system is at keeping computations on edge without making too many mistakes. When latency and accuracy are both important, this metric gives an overall measure of balance. Since lower thresholds have highest recall and that their precision is sufficiently high, F-1 score is also highest at lower thresholds as observed in Figure 17c.

Table 4. Average classification metrics across all thresholds for different calibration methods.

| Method | Precision | Recall | F1-score | Specificity | Accuracy |
|---|---|---|---|---|---|
| No Calibration | **0.8455** | 0.6142 | 0.6418 | 0.5420 | 0.6603 |
| DAC | 0.6854 | 0.3902 | 0.4168 | **0.7317** | 0.6058 |
| TS | 0.8305 | 0.7253 | 0.7341 | 0.4392 | 0.6953 |
| DC | 0.8227 | **0.7721** | **0.7655** | 0.4051 | **0.7163** |

*Specificity.* Indicates the proportion of cases where the edge model was not the ideal choice, and the system correctly decided to offload to the cloud. High specificity ensures that when the edge model is not ideal, the system correctly decides to offload to the cloud, improving accuracy. Conversely, low specificity means the system wrongly keeps tasks on edge even when the edge model is not ideal, hence reducing accuracy. Since higher thresholds will exclude almost all edge mispredictions, the highest specificity belongs to the highest threshold as also seen in Figure 17d.

*Accuracy.* Indicates the overall percentage of correct decisions, both keeping on edge when it's the ideal choice and moving to cloud when the edge model is not ideal. High accuracy means the system makes correct offloading decisions most of the time. However, accuracy alone can be misleading if the classes are imbalanced, e.g., if the majority of computations are always moved to the cloud. Similar to F-1 score, this metric is highest at lower thresholds as shown in Figure 17e.

To holistically compare all methods' classification metrics, Table 4 provide the average value for each metric over all thresholds. Among all methods, DC achieves the highest recall, F-1 score and accuracy. As we see in the next subsection, DC's superiority in classification metrics aligns closely with the trade-off results.

### 6.5 Accuracy and Performance Trade-off

To analyze HGN's accuracy and performance results, we consider deploying the cloud universal model on an NVIDIA A100 server GPU with a round-trip latency of 50*ms*, and an edge specialized model (Grasp-CLIP+) deployed on Jetson Orin NX 16GB in super mode. The latency of edge specialized model on different platforms are obtained from [15].

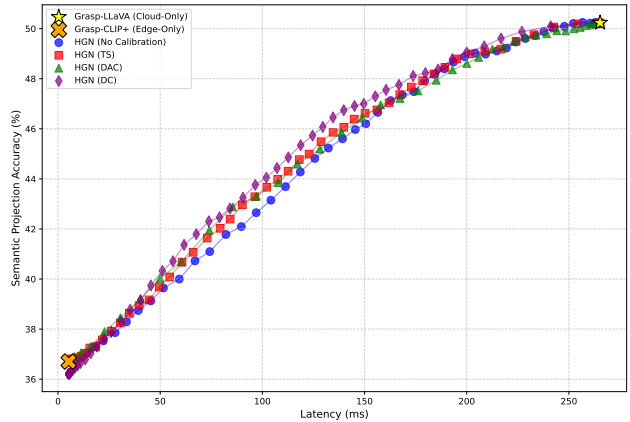

Fig. 18. HGN accuracy vs. performance trade-off with Orin NX 16GB as the specialized edge model.

Grasp-HGN: Grasping the Unexpected                                                                                    17Table 5. UUI for Implications Associated with Different Grasp Scenarios.

| # | Initial Grasp | Override Grasp | Description | Implications | UUI |
|---|---|---|---|---|---|
| S1 | Right | None or Right | $\hat{y}_E = y$ and $M = \begin{cases} E & \text{if conf} > \theta \\ C & \text{otherwise} \end{cases}$ | None | 0 |
| S2 | Wrong | Right | If $\hat{y}_E \neq y$, $\text{conf} > \theta$, and $\hat{y}_C = y$, then $M = C$ | Late | 1 |
| S3 | Wrong | None or Wrong (same misprediction) | $\hat{y}_E \neq y$ and $M = \begin{cases} E & \text{if conf} > \theta \\ C & \text{otherwise} \end{cases}$ | Incorrect | 5 |
| S4 | Wrong | Wrong (different misprediction) | If $\hat{y}_E \neq y$, $\hat{y}_C \neq \hat{y}_E$, and $\text{conf} < \theta$, then $M = C$ | Late and incorrect | 6 |
| S5 | Right | Wrong | If $\hat{y}_E = y$, $\text{conf} < \theta$, and $\hat{y}_C \neq y$, then $M = C$ | Late and incorrect override | 7 |

The accuracy vs. performance trade-offs for different thresholds are shown in Figure 18. As seen, HGN expands the Pareto front to include more options to choose from depending on the accuracy and performance requirements. Although the uncalibrated HGN already is already contributing to the Pareto front, calibrating HGN results in better Pareto fronts. Most notably, in line with the classification metric evaluations, HGN calibrated with DC consistently outperforms other methods. As an example, HGN can increase the semantic projection accuracy to 45% (+8.2% over the baseline edge model) while maintaining the average latency below a 150*ms* pre-shaping deadline.

From a user perspective, however, reasoning about latency vs. performance trade-off is nontrivial. Different applications might impose different tolerances for correctness and lateness of grasping. This motivates the need for a more structured metric that captures this trade-off in a user-centric way.

### 6.6 User Upsetness Index

In order to evaluate HGN from the user perspective, we propose User Upsetness Index (UUI). UUI provides a quantitative measure of user dissatisfaction by penalizing incorrect or delayed grasp decisions. While system-level metrics like accuracy and latency provide operational insights, UUI aims to reflect the actual user experience: whether the prosthesis performs correctly, and in a timely manner. As outlined in Table 5, a robotic prosthetic hand powered by HGN initially decides on a grasp ($\hat{y}_E$) using the specialized edge model ($E$) as shown in the first column, and depending on the prediction's confidence ($conf$) and the threshold used ($\theta$), the initial decision might be overridden ($\hat{y}_C$) by the universal cloud model ($C$). Here, the universal cloud model's grasp estimate overrides the specialized edge model's prediction only if they differ ($\hat{y}_E \neq \hat{y}_C$).

Based on the performance of the HGN, the user might experience different implications resulting from (i) correctness of the grasp used for the target object, or (ii) the timeliness of the grasping. Based on these two, the UUIs for each case is assigned. In doing so, although both correctness and latency of the grasping impacts the user experience, performing the correct grasp type is a more significant factor in user satisfaction. An incorrect grasp type prevents the robotic hand to grasp the target object, rendering the prosthesis non-functional. On the other hand, if the correct grasp type is chosen, even with increased latency, the prosthesis is still functional. We empirically selected a penalty of 5 for incorrect grasping and 1 for late but correct predictions. This ratio captures the intuition that a failed grasp is significantly more disruptive to the user than a delay, while still acknowledging the cost of latency. However, based on the application, tolerances may vary and their assignment requires a human subject experiment to provide the most relevant penalties tailored to the target population.

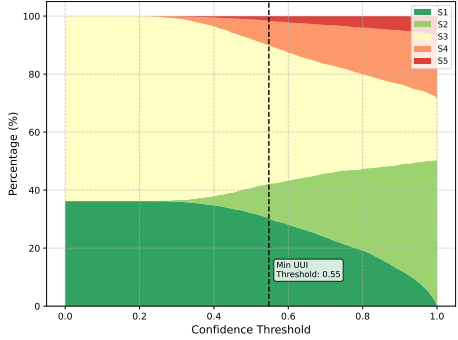
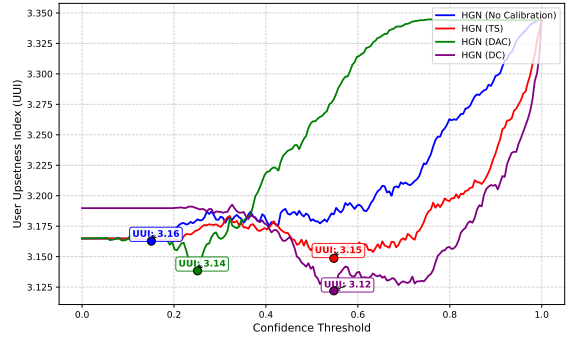

Fig. 19. Distribution of Grasp Scenarios Over Threshold for HGN (DC).

Fig. 20. User Upsetness Index vs. Threshold with Minimum Values.

The distribution of each scenario over varying thresholds is shown using a stack plot in Figure 19 for HGN with DC calibration. Starting from lower thresholds, we can observe only S1 and S3 scenarios occur. This is expected as with small thresholds, the edge model is always selected. For such cases, HGN has the least latency overhead at the price of no increased accuracy. However, as the threshold increases, both S1 and S3 scenarios occur less frequently and we see more of S2 and S4 scenarios. Although both scenarios increase the latency of grasping, we see more increase in S2 compared to S4, hence a positive net impact in the overall grasping correctness. Lastly, for larger thresholds, although the net impact on accuracy remains positive, due to higher reliance on the cloud model, the impact on latency is highly negative. Moreover, although small, S5 grows in proportion to threshold, further affecting the UUI.

The UUI analysis for all methods are depicted in Figure 20. Here, the lower the UUI is, the less the user will experience dissatisfaction. In line with previous findings, HGN with DC calibration results in the least UUI globally and for most thresholds. Other methods including HGN (DAC) and (DC) still outperform baseline HGN with no calibration. It is also noteworthy that DC provides a larger span of minimums over different thresholds, making it more robust for a potential system with a varying threshold.

With the above configurations, Grasp-HGN (DC) with a threshold of 0.55 achieves an overall accuracy of 42.3%, which is a 5.6% improvement over the edge model alone (36.7%). The average latency is 73*ms*, a 3.5× speedup over the standalone Grasp-LLaVA model. In terms of user experience, UUI drops from 3.16 (edge only) to 3.12. This is a relative 1.36% in UUI, corresponding to 19.1% of the total possible improvement within the observed UUI range (3.12 to 3.35).

### 6.7 Expected Real World Performance

The previous evaluations have focused on the semantic projection (unseen object, but known object type) to analyze most difficult case when all objects in the real-world are unknown. This marks the worst-case performance. To evaluate an expected performance, the percentage of objects of the seen type versus objects that demand semantic projection needs to be estimated. This requires reasoning about the dataset completeness versus object diversity in the real-world (beyond the scope of this article).

We postulate that a good dataset would represent 80% of the objects to interact with, and 20% are unknown demanding semantic projection. To evaluate performance of Grasp-HGN in this scenario, we use the entire validation set (i.e. 2000 images that the model has not seen but has seen similar objects) and 500 images from the test set (demanding semantic projection).

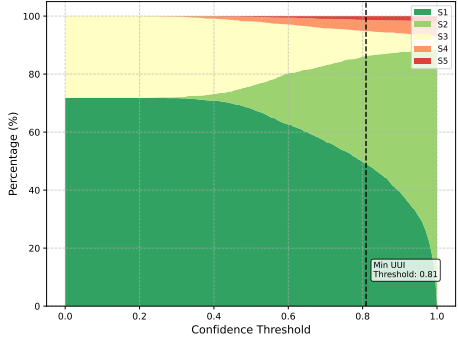 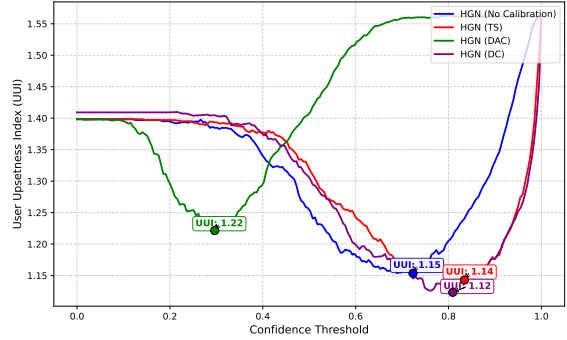

Fig. 21. Grasp Scenarios Over Threshold for HGN (DC) for 80/20 mix.

Fig. 22. User Upsetness Index vs. Threshold for 80/20 mix.

Figure 22 measures user upsetness over threshold and calibration method. Overall user upsetness starts out much lower at 1.4 as the set now includes images with already seen object types. Now, DC with 0.81 threshold yields the lowest UUI of 1.12. Which is also our final chosen configuration.

Figure 21 shows the detailed distribution over scenarios. Accuracy reaches 86.0% with the selected 0.81 threshold. Note that higher thresholds further increase accuracy (S1 + S2). However, the very upsetting scenarios (S4 +S5) increase as well which results in increasing upsetness.

Overall, our Grasp-HGN (DC) with a threshold of 0.81 reaches an average accuracy of 86% which is a 12.2% improvement over the edge model alone (71.3%). Grasp-HGN achieves 97.2% of Grasp-LLaVA's accuracy (88.4%). UUI drops from 1.4 (edge only) to 1.12. The average latency is 117.8*ms*, a 2.2× speedup over Grasp-LLaVA.

## 7 Discussion

This publication lays the foundations for reasoning about the lab to real-world gap by using semantic projection. Grasp-HGN highlights the opportunities of edge / cloud hybrid networks to achieve latency and accuracy. The User Upsetness Index provides a framework to reason about multi-objective problems from a user perspective. Our work lays the foundations for additional research toward a deployable product.

Deployment optimizations such as, model distillation, quantization, pruning, or other compression can further reduce the latency (both of edge and cloud models). This is a heavily researched topic, especially in the context of LLMs. Deployment optimizations can be considered in addition to the hybrid model proposed here. While an edge-only model is preferred, e.g. due to lower complexity, latency, privacy, even with aggressive deployment optimizations limitations persist. As such, we analyze deployment of LLaVA using performance numbers from NVIDIA which includes heavy optimizations such as INT4 precision in super mode. Yet, A too large, heavy and power consuming compute device causes practical disadvantages that would drive the user upsetness index up much more than the LLMs zero shot capability can reduce it. Optimizations can shift the boundary but don't eliminate it. Some deployment/application/use-case scenarios will necessitate an edge/cloud collaboration. Additional requirements that might favor cloud involvement could include greater model flexibility (experts of models), online adaptive learning, or a shared learning approach. Nonetheless, our future work also aims at pushing the edge boundaries further by designing specialized edge models that aim to achieve the semantic projection capability of an LLM but at edge cost.

This paper enables further research to push for increased semantic projection capability at the edge. By leveraging large language models (LLMs), it establishes a reference point. This benchmark is critical for evaluating the effectiveness of future lightweight alternatives intended for real-time systems.

Deployment of cloud-assisted systems, however, introduce naturally additional concerns, such as network latency sensitivity, privacy and security. Network latency and connectivity can impact system responsiveness, particularly in time-sensitive applications such as prosthetic control. A key to mitigate its impact is to impose a maximum response deadline after which a cloud response is ignored. In case of network failure (or exceeding delay) the system behaves as local (as our system enacts the local prediction first). Privacy concerns arise when transmitting visual data to external servers. To mitigate privacy issues, HGN supports transmitting the locally computed CLIP feature vector instead of raw images for the cloud LLM. This is feasible as both the edge and the cloud model rely on CLIP for image feature extraction.

This architecture provides a foundation for flexible deployment strategies based on task complexity, latency requirements, and resource availability. In scenarios where the edge is sufficient, full local inference ensures responsiveness and reliability. For tasks requiring greater generalization or contextual reasoning, selective cloud augmentation becomes a viable option. The edge-cloud partitioning is therefore a core system design decision.

Future work will explore adaptive strategies for dynamic task partitioning, efficient handling of intermittent connectivity, and cloud-based online adaptation. These directions will further refine the edge-cloud balance and advance the practical deployment of semantic projection models in real-world assistive and robotic systems.

## 8 Conclusion

Robotic prosthetic hands offer functional solutions for amputees, but their effectiveness in real-world scenarios is hindered by challenges in capturing semantic projection, and computational efficiency. We introduced Semantic Projection a concept to evaluate the ability of models to generalize to unseen objects with the same grasp type and proposed Grasp-LLaVA, reimagining LLaVA by incorporating text-based reasoning to achieve semantic projection. Current SOTA approaches for visual grasp estimation, including classification and detection-based methods, fail to perform adequately on unseen objects, achieving only 15.3-36.7% grasp accuracy on such objects. On the other hand, we demonstrated that Grasp-LLaVA achieves 50.2% accuracy on unseen object types, outperforming SOTA. Additionally, we showed that by deploying a well-calibrated edge specialized model and Grasp-LLaVA on cloud in HGN, the Pareto front is expanded with HGN (DC) providing the best in latency vs. accuracy trade-off. Lastly, we proposed and measured User Upsetness Index (UUI) to evaluate HGN based on how the user experiences grasping, taking into account correctness and lateness of the grasp being executed. We showed HGN (DC) results in the least UUI globally while provides a larger span of minimums over different thresholds, making it more robust for a potential system with a varying threshold. As such, we showed that HGN (DC) with Threshold=0.81 achieves an overall accuracy of 86%, an average latency of 117.8$ms$, and a user upsetness index of $UUI = 1.12$, improving average accuracy, latency and UUI.

## Acknowledgments

We thank Mohammadreza Taesiri for his support and insightful guidance in working with LLaVA. We also gratefully acknowledge Professor David Kaeli for providing access to A100 GPUs, which were essential for training and large-scale experimentation.